\documentclass[10pt,twocolumn,letterpaper]{article} 

\usepackage{avss}
\usepackage{times}
\usepackage{epsfig}
\usepackage{graphicx}
\usepackage{amsmath}
\usepackage{amssymb}
\usepackage{url}
\usepackage[ruled]{algorithm2e}
\usepackage{bm}
\usepackage{subfigure}
\usepackage{fancyhdr}

\avssfinalcopy 
\def\avssPaperID{9} 

\newcommand{\reffig}[1]{Figure \ref{#1}}

\begin{document}

\title{Latent Embeddings for Collective Activity Recognition}

\author{Yongyi Tang$^1$, Peizhen Zhang$^2$, Jian-Fang Hu$^2$ and Wei-Shi Zheng$^2$$^3$\thanks{Corresponding author.}\\
$^1$School of Electronics and Information Technology, Sun Yat-Sen University, China\\
$^2$School of Data and Computer Science, Sun Yat-Sen University, China\\
$^3$The Key Laboratory of Machine Intelligence and Advanced Computing \\
(Sun Yat-sen University), Ministry of Education, China\\
{\tt\small \{tangyy8,zhangpzh5\}@mail2.sysu.edu.cn, hujf5@mail.sysu.edu.cn, wszheng@ieee.org}
}

\maketitle
\thispagestyle{fancy}
\pagenumbering{gobble}
\lfoot{978-1-5386-2939-0/17/\$31.00 \copyright 2017 IEEE}

\begin{abstract}
Rather than simply recognizing the action of a person individually, collective activity recognition aims to find out what a group of people is acting in a collective scene.
Previous state-of-the-art methods using hand-crafted potentials in conventional graphical model which can only define a limited range of relations.
Thus, the complex structural dependencies among individuals involved in a collective scenario cannot be fully modeled. In this paper, we overcome these limitations by embedding latent variables into feature space and learning the feature mapping functions in a deep learning framework. The embeddings of latent variables build a global relation containing person-group interactions and richer contextual information by jointly modeling broader range of individuals. Besides, we assemble attention mechanism during embedding for achieving more compact representations.
We evaluate our method on three collective activity datasets, where we contribute a much larger dataset in this work. The proposed model has achieved clearly better performance as compared to the state-of-the-art methods in our experiments.
\end{abstract}

\section{Introduction}

Recognizing what a group of people is doing, which is named the collective activity, is critical and useful in some real-world applications including visual surveillance. A critical point for collective activity analysis is to model the interaction between persons in the collective scenario, and inferring relations among individuals in images/videos remains challenging.

Existing approaches for the collective activity recognition typically modeled the collective interactions in terms of person-person interactions \cite{choi2014understanding, lan2012discriminative, amer2014hirf,hajimirsadeghi2015learning}. For instance, Lan \etal \cite{lan2010beyond} explicitly modeled pairwise potential between individuals based on atomic action labels. Choi \etal \cite{choi2014understanding} explored several hand-crafted interaction features for constructing pairwise potentials. Chang \etal \cite{chang2015learning} chose to model person-person interaction in collective scene by an interaction metric matrix.
In addition, deep learning models such as recurrent neural network are also proposed for modeling pairwise person interaction \cite{deng2015deep, ibrahim2015hierarchical, deng2015structure}.
 The person-person interaction based models intend to describe activities from a local perspective, but it causes ambiguities. Besides, those models are intrinsically limited in capturing high-level collective activities due to the inherent visual ambiguity caused by the activity
 invaders\footnote{We use activity invaders to indicate the individuals irrelevant to the activity.} and local pattern uncertainty.

In this work, we aim to describe collective interactions from a more global perspective, where the interactions between each anchor individual and the rest individuals are explicitly modeled. We call this interaction the \emph{person-group interaction}.
For effectively capturing person-group interactions, we introduce a set of latent variables that are modeled by jointly considering all the related person in a collective scenario. We infer those latent variables with complicated dependencies by embedding them into feature space using deep neural network instead of defining hand-crafted potentials in conventional graphical model. The benefits are twofold. First, by utilizing embedding-based method, our model is able to model more complex collective structures beyond pairwise person-person structures. Second, the non-linear dependencies between person and group can be inferred by discriminative learning procedure in deep learning framework. To obtain a more concise collective activity representation, an attention mechanism is employed to modify the contextual structure by setting different impact factors for each individuals during the embedding procedure.

In summary, the contributions of our paper are threefold. Firstly, a latent variable model
capable of capturing complex connections among individuals is developed for collective activity recognition.
Secondly, the complicated dependencies between person and group are represented by latent variable embedding and an attention mechanism is integrated for obtaining a compact embedding representation.
Thirdly, a new dataset with more activity samples is collected for the benchmarking of collective activity recognition.

\begin{figure}[t]
\begin{center}
   \includegraphics[width=0.8\linewidth]{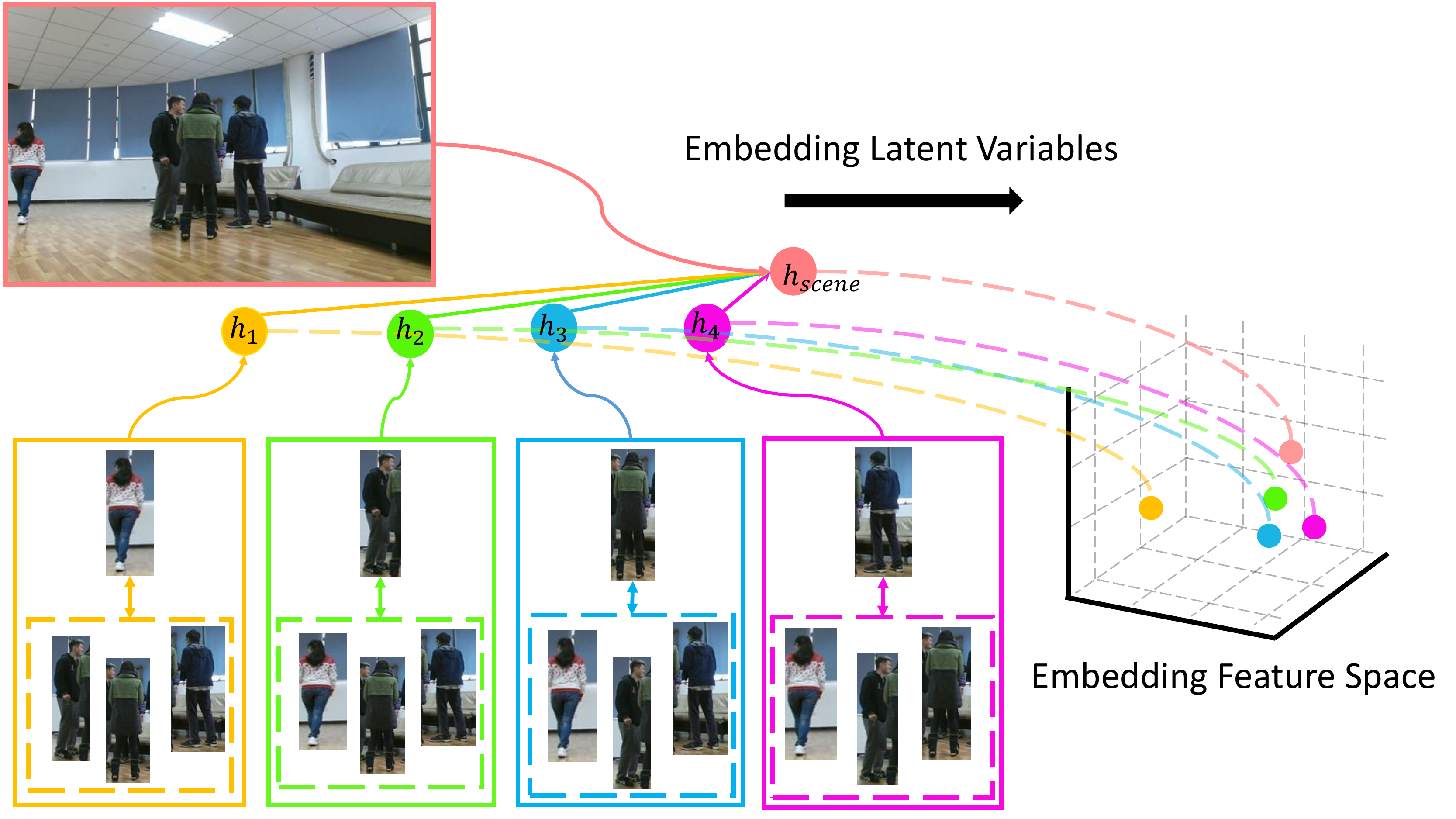}
\end{center}
   \centering\caption{Illustration of constructing latent variables to model person-group interaction. The latent variable $h_i$ captures local person-group interaction of person $i$ while $h_{scene}$ mines the global interaction by aggregate all the local interaction information. To effectively model complex dependencies, we learn the representation of latent variables in embedding feature space.}
\label{fig:fig2}
\end{figure}

\begin{figure*}[t]
\begin{center}
   \includegraphics[width=0.9\linewidth]{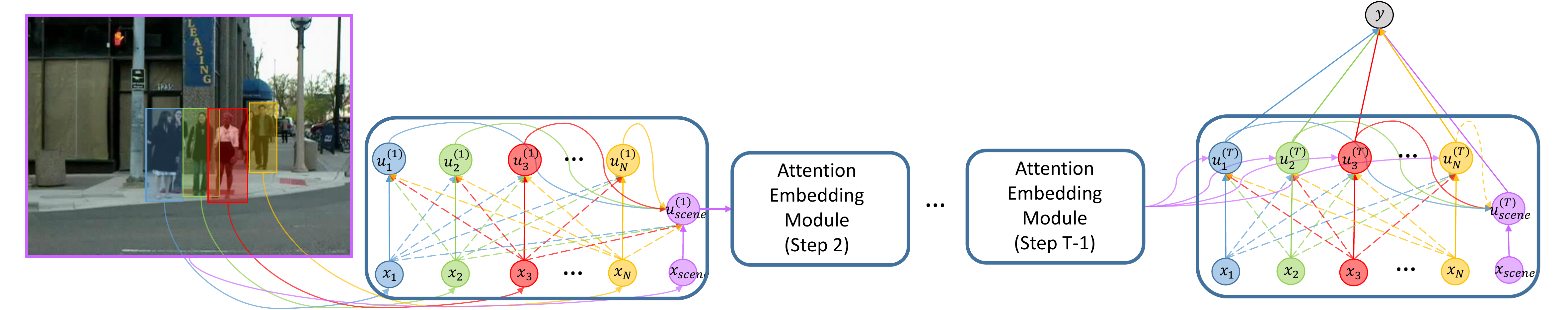}
\end{center}
   \centering\caption{The pipeline of our Latent Variable Embedding procedure. We represent the sub-procedure at each time step as Attention Embedding Module for simplicity. The output of module in time step $t$ is $\mathbf{u}_{scene}^{(t)}$ which indicates the summarized person-group interaction in collective scenario. Then the interaction information is propagated in an iterative fashion. Note that the dash line between $\mathbf{u}_N^{(T)}$ and $\mathbf{u}_{scene}^{(T)}$ means that the interaction of person in yellow barely related to the current collective activity. Activity classification performs at the last time step $T$.}
\label{fig:fig3}
\end{figure*}
\section{Our Approach}
Instead of capturing collective structures using person-person interaction only, we consider mining more globally structural interactions between each individual and the rest group (other individuals).
Here, we utilize latent variables to capture the complicated dependencies between person and group in collective activity scenario. Rather than directly infer the latent states, we exploit the embeddings of latent variables parametrized by a deep neural network to represent structural information from a global view and then explicitly model person-group interaction for collective activity recognition.

\thispagestyle{empty}
\subsection{Modeling Collective Activity with Latent Variable}
In this section, we aim to construct a mid-level feature representation indicating the collective interactions among individuals via latent variables encompassed by a graphical model.
For simplification, we denote $x_i$ as visible variable of person $i$, where $i\in\mathcal{V}_p$ and $\mathcal{V}_p$ is the set of people involved in a collective scene, and the visible variable of scene as $x_{scene}$. In addition, we use $h_i$ and $h_{scene}$ to indicate the hidden variables of the $i-th$ individual and scene, respectively.
Based on the isolated representation of entities, the interactions among actors, related group and the context can then be captured by the corresponding latent variables.
Figure \ref{fig:fig2} provides a graphical representation of our model.

Thus, the posterior probability of each latent variable can be expressed as $p(h_i|x_i,{\{x_j\}}_{j\in\mathcal{V}_p\backslash i},h_{scene})$ and $p(h_{scene}|x_{scene},{\{x_i\}}_{i\in\mathcal{V}_p},{\{h_i\}}_{i\in\mathcal{V}_p})$. It means that the hidden variable $h_i$ captures the person-group interaction information for anchor person $i$ and $h_{scene}$ captures all the interaction in the collective scenario from global view. Building upon the latent variables, the collective activities can be recognized by jointly considering local person-group interactions and global context as $p(y|{\{h_i\}}_{i\in\mathcal{V}_p},h_{scene})$.

\subsection{Latent Variable Embedding}
However, even though we can define the posterior probability of these latent variables, the exact inference procedure is difficult and sometimes even intractable in conventional graphical model based on hand-crafted potentials.
Inspired by \cite{dai2016discriminative} where latent variables are embedded into feature space for structural modeling, we utilize a deep neural network to capture the non-linear dependencies among person-group interactions and represent it as the embeddings of latent variables.
The embeddings can be viewed as an indication of the posterior probabilities.

As shown in \reffig{fig:fig3}, we develop a mean-field like procedure in order to approximate inference and capture the person-group interaction during embedding. Thus, the embeddings of latent variable can be learnt in the iterative manner introduced below.

We first denote $\mathbf{u}_i^{(t)}$ as the embedding of latent variable $h_i$ and formulate it by jointly considering the unary image feature $\mathbf{x}_i$ of person $i$, the averaged appearance feature $(\sum_{j\in\mathcal{N}(i)}\mathbf{x}_j)/|\mathcal{N}(i)|$ of all the neighbours of person $i$, and the embedding of global scene from last step $\mathbf{u}_{scene}^{(t-1)}$. We denote the neighbouring persons of $i$ as $\mathcal{N}(i)$, $\mathcal{N}(i)\subseteq\mathcal{V}_p$. Then, the update of $\mathbf{u}_i^{(t)}$ is below: $\forall i\in\mathcal{V}_p$,
\begin{equation} \label{eq:u_i}
\begin{split}
\mathbf{u}_i^{(t)} &=(1-\lambda)\cdot\mathbf{u}_i^{(t-1)} \\
&+\lambda\cdot\sigma(\mathbf{W}_u [\mathbf{x}_i; \frac{\sum_{j\in\mathcal{N}(i)}\mathbf{x}_j}{|\mathcal{N}(i)|}; \mathbf{u}_{scene}^{(t-1)}]),
\end{split}
\end{equation}
where "$;$" indicates vector vertical concatenation, $\sigma(\cdot)$ is a rectified linear unit and $\lambda$ is the update step size. Here, we omit the biases term for simplicity. Intuitively, the aggregated neighbour feature $(\sum_{j\in\mathcal{N}(i)}\mathbf{x}_j)/|\mathcal{N}(i)|$ is employed to represent group appearance information, while $\mathbf{u}_{scene}^{(t-1)}$ indicates the global context information. Thus, the local person-group interaction can be represented by the embedding $\mathbf{u}_i^{(t)}$.

Likewise, $\mathbf{u}_{scene}^{(t)}$ is the embedding of $h_{scene}$, and it aims to capture collective interactions from a global view. To this end, we formulate it by the global image feature $\mathbf{x}_{scene}$, the pooled low-level representation of person, \ie $\sum_{i\in\mathcal{V}_p}\mathbf{x}_i/|\mathcal{V}_p|$ and the aggregate embeddings of individuals, \ie $\sum_{i\in\mathcal{V}_p}\mathbf{u}_i^{(t)}/|\mathcal{V}_p|$. Specifically, it can be formulated as:
\begin{equation} \label{eq:u_s}
\begin{split}
\mathbf{u}_{scene}^{(t)} &= (1-\lambda)\cdot\mathbf{u}_{scene}^{(t-1)} \\
&+\lambda\cdot\sigma(\mathbf{W}_{s} [\mathbf{x}_{scene}; \frac{\sum_{i\in\mathcal{V}_p}\mathbf{x}_i}{|\mathcal{V}_p|}; \frac{\sum_{i\in\mathcal{V}_p}\mathbf{u}_i^{(t)}}{|\mathcal{V}_p|}]),
\end{split}
\end{equation}
Thus, $\mathbf{u}_{scene}^{(t)}$ can be considered as global relation representation since it models the non-linear dependencies of individuals and their local relations.

Based on the embeddings of latent variable, we can define the posterior probability of assigning activity label $\mathbf{y}$ to a given sample by non-linearly combining all the embeddings together:
\begin{equation}\label{eq:p_y}
\begin{split}
p(\mathbf{y}&|{\{\mathbf{u}_i\}}_{i\in\mathcal{V}_p},\mathbf{u}_{scene}) = \\
&\phi(\mathbf{W}_{out}\sigma(\mathbf{W}_{y}[\frac{\sum_{i\in\mathcal{V}_p}\mathbf{u}_i^{(T)}}{|\mathcal{V}_p|};\mathbf{u}_{scene}^{(T)}])).
\end{split}
\end{equation}
Here, $\phi$ is an activation function used for scaling the network outputs and we set it as softmax.

Finally, we use the following cross entropy loss function to measure the consistency between the model outputs and manual annotations:
\begin{equation}\label{eq:loss}
\begin{split}
L(\theta)&=-\sum_{k=1}^{\mathrm{K}}y_k\log(p(y_k|{\{\mathbf{u}_i^{(T)}\}}_{i\in\mathcal{V}_p},\mathbf{u}_{scene}^{(T)})),
\end{split}
\end{equation}
where $\theta$ is the model parameters to be learned, $K$ is the number of collective activity labels and $y_k$ is $1$ if the frame belongs to class $k$ and $0$ otherwise. 
The model parameters are optimized using the back propagation through time (BPTT) algorithm. 

\subsection{Embedding with Attention}
\thispagestyle{empty}
\label{sec:Attention}

Note that the update of $\mathbf{u}_{scene}^{(t)}$ in Eq.\eqref{eq:u_s} involves the summation of individual embeddings ${\{\mathbf{u}_i^{(t)}\}}_{i\in\mathcal{V}_p}$, which means that all the person-group interactions are equally connected to the group activity. However, to correctly discover the collective structure information, one should pay more attention to some relevant person-group interactions. For example, in a waiting scenario presented in \reffig{fig:fig3}, those individuals who are waiting in line should be paid more attention to, since their person-group interactions are strongly relating to the activity, while the subjects walking behind are less valuable for the recognition, and sometimes even cause ambiguity.
Thus the influence of the interactions between walking subjects and waiting group should be suppressed in this case. Inspired by the recent success of attention models for sequential modeling \cite{chorowski2015attention, ramanathan2015detecting}, we use an attention mechanism to encode the relevance of each individual embedding and scene embedding as:
\begin{equation}\label{eq:alpha}
\alpha_i^{(t)}=tanh(\mathbf{w}_g^\mathrm{T}\mathbf{u}_i^{(t)} + \mathbf{w}_{gs}^\mathrm{T}\mathbf{u}_{scene}^{(t-1)}),
\end{equation}
where $\mathbf{w}_g,\mathbf{w}_{gs}\in\mathbb{R}^d$.
Given the relevance of individuals in the collective scenario, we can measure the importance of the person-group interactions derived from individual $i$ as:
\begin{equation}\label{eq:gi}
g_i^{(t)}=\frac{e^{\alpha_i^{(t)}/\tau}}{\sum_{j\in\mathcal{V}_p}e^{\alpha_j^{(t)}/\tau}},
\end{equation}
where $\tau$ is the softmax temperature parameter.

By considering all the individuals in the given collective scenario together,
we can reformulate the embedding of scene as following:
\begin{equation}\label{eq:newus}
\begin{split}
\mathbf{u}_{scene}^{(t)} &= (1-\lambda)\cdot\mathbf{u}_{scene}^{(t-1)} \\
&+\lambda\cdot\sigma(\mathbf{W}_{s} [\mathbf{x}_{scene}; \frac{\sum_{i\in\mathcal{V}_p}\mathbf{x}_i}{|\mathcal{V}_p|}; \sum_{i\in\mathcal{V}_p}g_i^{(t)}\mathbf{u}_i^{(t)}).
\end{split}
\end{equation}

\section{Experimental Results}

For evaluation, we tested our model on three collective activity datasets: collective activity dataset \cite{Choi_VSWS_2009}, the collective activity extended dataset and a newly proposed dataset by ourselves denoted as CA Dataset, CAE Dataset and SYSU-CA Dataset, respectively. We have compared our model with the state-of-the-art collective activity recognition methods \cite{antic2014learning, lan2012discriminative, choi_eccv12, hajimirsadeghi2015learning, hajimirsadeghi2015visual, deng2015deep, deng2015structure, ibrahim2015hierarchical}. In the following, we first provide some implementation details and then report our results on these three benchmark.

For feature representation, we used the feature maps obtained in the ``pool5'' layer of two-stream ResNet-50 net (pretrained on the UCF101 action set \cite{twostream01}) as our two-stream feature. For each person in the collective scenario, we extracted its two-stream feature as our individual feature. We also extracted the two-stream feature from the entire collective image as the feature representation of a scene. Our algorithm was implemented using the Tensorflow package \cite{tensorflow2015-whitepaper}. We empirically set the Softmax temperature parameter $\tau$ and the update step size as 0.25 and 0.3 respectively. The hidden units of latent embedding was set as 256. The dropout weight of the dropout layer employed in Eq.(\ref{eq:p_y}) was set to 0.5 during training phase.
We deployed Xavier Initiaizer suggested in \cite{glorot2010understanding} and optimized parameters with Adam Optimization strategy \cite{kingma2014adam}.

We also conducted two baselines including Image Classification and Person Classification for comparison. In Image Classification, we built a softmax classifier on top of two-stream feature of each single frame. While in Person Classification, we constructed a feature representation by averaging features over all people instead.

\subsection{Collective Activity (CA) Dataset}
\thispagestyle{empty}

The collective activity dataset contains 44 video clips of 5 collective activities including crossing, waiting, queueing, walking and talking. Each participant appeared in the videos was annotated every 10 frames with a bounding box and the collective activity labels were provided for evaluation. We followed exactly the testing protocol used in \cite{deng2015structure} and compared our method with the state-of-the-art methods in Table \ref{Tab:CompResCAD}. We set the model parameters $T$ variables as 3. Their effects will be further discussed in section \ref{sec:moreEvaluation}.

As shown, our model outperformed both the deep learning based and non-deep learning based competitors, and obtained the state-of-the-art results on the Collective Activity Dataset. Specifically, our method achieved an accuracy of 85.4\%, which is about 2\% higher than that of Cardinality Kernel model. Our model outperformed the Deep Structure Model by a margin of 4\%. The results demonstrate that our proposed person-group interaction based modeling performs better than the existing person-person interaction based modelings.

\begin{figure*}[]
	\begin{center}
	\resizebox{.75\textwidth}{!}{
		\includegraphics[width=.951\linewidth]{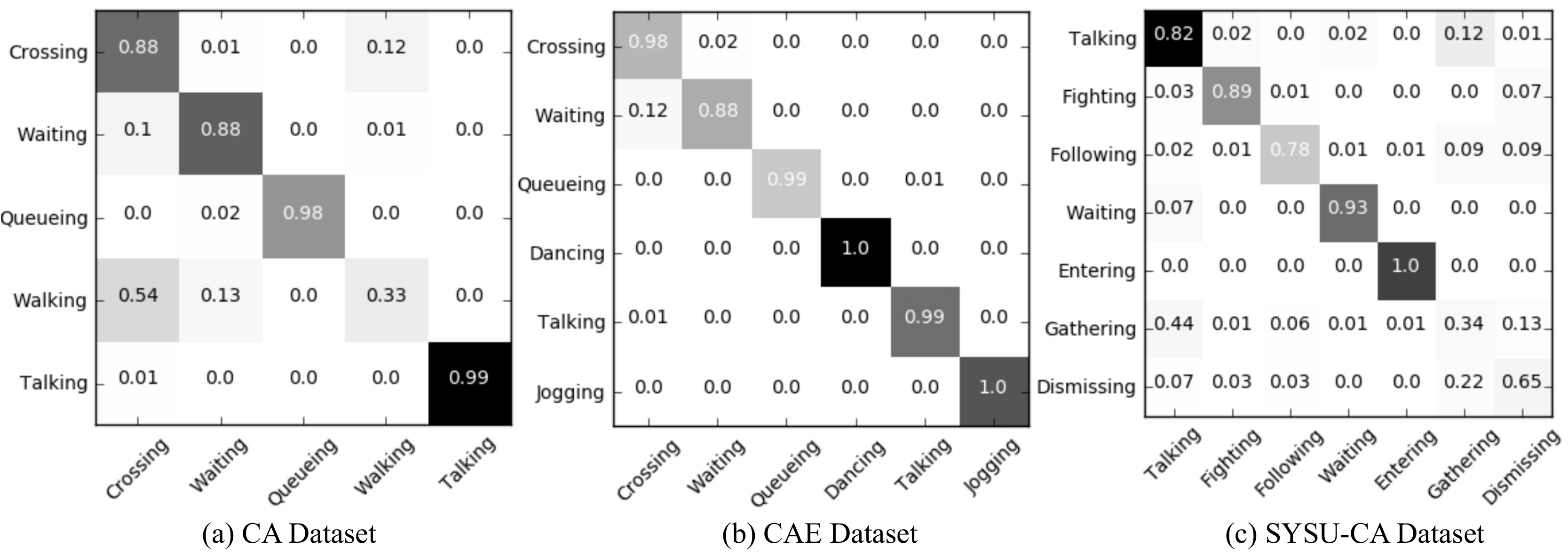}
		}
\end{center}		
	\caption{Confusion matrices obtained by our method.}
	\label{Fig:confusion}
\end{figure*}

The relevant confusion matrix obtained by our method is presented in Figure \ref{Fig:confusion} (a). It reveals that our method can achieve a good result for the recognition of activities like talking, waiting, and queueing . We also observe that our method often misclassified activity Walking as Crossing. This is because that subjects in both walking and crossing actvities performed similar atomic action (walking), and the person-group interactions in these activities were not as distinguishable as those in the other activities such as talking and queueing. This result is consistent with the claim drawn in \cite{Choi_CVPR_2011} that the walking activity in this set could be biased.

\begin{table}[]
	\begin{center}
	\resizebox{.45\textwidth}{!}{
		\begin{tabular}{|l|c|}
			\hline
			Method & Accuracy \\
			\hline\hline
			Image Classification (Two-stream feature + softmax) &  71.2\%\\
			\hline
			Person Classification (Average two-stream features + softmax) & 77.2\%\\
			\hline\hline
			Latent Constituent Model \cite{antic2014learning} & 75.1\% \\
			\hline
			Discriminative Latent SVM Model \cite{lan2012discriminative} & 79.7\% \\
			\hline
			Unified Tracking and Recognition \cite{choi_eccv12} & 80.6\% \\
			\hline
			HCRF-Boost \cite{hajimirsadeghi2015learning} & 82.5\% \\
			\hline
			Cardinality Kernel \cite{hajimirsadeghi2015visual} & 83.4\% \\
			\hline
			Deep Structure Model \cite{deng2015deep} & 80.6\% \\
			\hline
			Sructure Inference Machines \cite{deng2015structure} & 81.2\% \\
			\hline
			Hierarchical Deep Temporal Model \cite{ibrahim2015hierarchical} & 81.5\% \\
			\hline
			Our Model & {\bf 85.4\%} \\
			\hline
		\end{tabular}
		}
	\end{center}	
	\caption{Comparison on the Collective Activity Dataset.}
	\label{Tab:CompResCAD}
\end{table}

\subsection{Collective Activity Extended (CAE) Dataset}

By replacing the walking activity with two activities dancing and jogging, the Collective Activity extended Dataset with 6 collective activities was proposed in \cite{Choi_CVPR_2011}. For evaluation, we also set $T$ as 3 and followed the evaluation protocol in \cite{deng2015structure}.
Table \ref{Tab:CompResCADE} presents the detailed comparison results. As shown, our model can obtain an accuracy of 97.94\%, which is 7\% superior to the best result obtained by the Structure Inference Machines model \cite{deng2015structure}. This again demonstrates the effectiveness of the proposed person-group interaction modeling for collective activity recognition.

By exactly examining the confusion table obtained by our method in Figure \ref{Fig:confusion} (b), our model can obtain good recognition results for most of the activities. We also observe that about 12\% of the Waiting samples were misclassified as Crossing, since in most of misclassified scenarios, waiting activity was usually followed by crossing, and the transition boundary between these two activities is indistinguishable, which are usually labelled as crossing.

\begin{table}[]
	\begin{center}
	\resizebox{.45\textwidth}{!}{
		\begin{tabular}{|l|c|}
			\hline
			Method & Accuracy \\
			\hline\hline
			Image Classification (Two-stream feature + softmax) &  92.26\%\\
			\hline
			Person Classification (Average two-stream features + softmax) & 95.10\%\\
			\hline\hline
			CRF + CNN \cite{deng2015structure} & 86.75\% \\
			\hline
			Structural SVM + CNN \cite{deng2015structure} & 87.34\% \\
			\hline
			Sructure Inference Machines \cite{deng2015structure} & 90.23\% \\
			\hline
			Our Model & {\bf 97.94\%} \\
			\hline
		\end{tabular}
		}
	\end{center}
	\caption{Comparison on the Collective Activity Extended Dataset.}
	\label{Tab:CompResCADE}	
\end{table}

\subsection{SYSU Collective Activity (SYSU-CA) Dataset}

For more in-depth evaluation, we also collected a new multi-view collective activity dataset. 
This dataset includes 7 different collective activities (\emph{Talking}, \emph{Fighting}, \emph{Following}, \emph{Waiting}, \emph{Entering}, \emph{Gathering} and \emph{Dismissing}) distributed in total 285 video clips which were captured from 3 different views.
Compared with other existing datasets, this set is unique in the following aspects:
1) each activity was captured from three different views; 2) the set contains more activity samples for collective activity analysis; 3) the dynamic motions in collective activities are more complex.

Our results are obtained by different methods on this dataset followed in four different settings: view1, view2, view3 and an integrated version. As for single view evaluation, we employed three-fold cross validation protocol, where two-thirds of the videos from the corresponding view were used for training and the rest for testing. In the integrated evaluation setting, we report the accumulative accuracies obtained on separate views. In details, we set the parameters $T$ as 4, respectively.

The experimental results are presented in Table \ref{Tab:CompResNewCAD} and Figure \ref{Fig:confusion} (c). Compared with the baselines Image Classification and Person Classification, our method achieved the best recognition result on most of the view settings and obtained an accuracy of 85.85\% on the total setting. We also observe that our baseline Image Classification model can obtain a reliable performance on this set, while the Person Classification performed unsatisfactorily. We further concluded that our model is robust to view variation since the results on three different views were consistent and satisfactory. However, since the person-group interactions were explicitly modeled, the performance has been further improved. The confusion table in Figure \ref{Fig:confusion} (c) indicates that our method often confuses activities Gathering and Dismissing with each other.
This can be attributed to that the individuals in both activities had high similarity on the spatial and temporal distribution.

\begin{table}[]
\begin{center}
\resizebox{.45\textwidth}{!}{
\begin{tabular}{|l|c|c|c|c|}
\hline
Method & View1 & View2 & View3 & Total \\
\hline\hline
Image Classification & 86.89\% & 81.62\% & 82.60\% & 83.70\%\\
\hline
Person Classification & 63.90\% & 64.42\% & 66.38\% & 64.90\%\\
\hline
Ours & 85.58\% & 87.02\% & 84.92\% & {\bf 85.84\%}\\
\hline
\end{tabular}
}
\end{center}
\centering\caption{Comparison on the SYSU Collective Activity Dataset.}
\label{Tab:CompResNewCAD}
\end{table}

\subsection{More Discussions}

\begin{table}[]
\begin{center}
\resizebox{.47\textwidth}{!}{
\begin{tabular}{|l|c|c|c|c|c|}
\hline
Dataset & $T$=1 & $T$=2 & $T$=3 & $T$=4 & $T$=15\\
\hline\hline
CA Dataset & 84.51\% & {\bf 85.45\%} & {\bf 85.45\%}  & 84.89\% & 82.63\% \\
\hline
CAE Dataset & 97.26\% & 96.47\% & {\bf 97.94\%} & 97.55\% & 97.06\%\\
\hline
SYSU-CA Dataset & 85.23\% & 82.96\% & 85.79\% & {\bf 85.84\%} & 80.99\%\\
\hline
\end{tabular}
}
\end{center}
\caption{Evaluation on iteration step number $T$.}
	\label{Tab:EvalT}
\end{table}

\label{sec:moreEvaluation}

\begin{table}[]
\begin{center}
\resizebox{.45\textwidth}{!}{
\begin{tabular}{|l|c|c|}
\hline
Dataset & Without Attention & With Attention\\
\hline\hline
CA Dataset & 83.68\% & {\bf 85.45\%} \\
\hline
CAE Dataset & 97.45\% & {\bf 97.94\%}  \\
\hline
SYSU-CA Dataset& 85.20\% & {\bf 85.79\%} \\
\hline
\end{tabular}
}
\end{center}
\caption{Evaluation on the Attention mechanism.}
	\label{Tab:EvalAtt}
\end{table}

\thispagestyle{empty}
\noindent
\textbf{Efficiency of our model.} Compared with baseline models, the difference in our model is that we explicitly model person-group interaction in latent space so that our model is able to compensate the individual information with group context such that our model outperforms most of the baseline models with the same feature in all dataset as shown in Table \ref{Tab:CompResCAD}, Table \ref{Tab:CompResCADE} and Table \ref{Tab:CompResNewCAD}.

\noindent
\textbf{Effect of iteration step $T$.} Table \ref{Tab:EvalT} provides the results of varying the iteration step number $T$ in our embedding procedure. In this experiment, attention mechanism was employed and the number of hidden neurons was set as 256. We can observe that, a better recognition result can be obtained by setting $T$ as 3 or 4 in most of the cases, which means that the collective interactions can be effectively discovered by our embedding model with a quite small $T$.

\noindent
\textbf{With vs. without attention.}  Here, we investigated the effect of the employed attention embedding mechanism. For comparison, we set the parameters $T$ and number of  hidden units as 3 and 256, respectively. The detailed comparison results are presented in Table \ref{Tab:EvalAtt}. As shown, using attention mechanism can always benefit the recognition. Especially on the collective activity dataset, the introduced attention mechanism can improve the accuracy by a margin of 2\%, which demonstrates that the attention mechanism can help to suppress the influence of activity invaders and thus obtained a better activity representation.

\section{Conclusion}
In this paper, we developed a latent embedding model for collective activity recognition. By embedding the latent variables in the collective graphical model and combining with an attention mechanism, our method can effectively capture the complex collective structures depicted in collective activity videos (images) and obtain the state-of-the-art results on two benchmarking datasets and a new collective activity set.

\thispagestyle{empty}
\section*{Acknowledgment}
This work was supported partially by the National Key Research and Development Program of China (2016YFB1001002, 2016YFB1001003), NSFC (No.61522115, 61472456, 61628212), Guangdong Natural Science Funds for Distinguished Young Scholar under Grant S2013050014265, the Guangdong Program (No.2015B010105005), the Guangdong Science and Technology Planning Project (No.2016A010102012, 2014B010118003), and Guangdong Program for Support of Top-notch Young Professionals (No.2014TQ01X779).

{\small
\bibliographystyle{ieee}
\bibliography{egbib}
}

\end{document}